\documentclass[sigconf]{acmart}
\usepackage{makecell}
\usepackage{tabularx}  
\usepackage{booktabs}  
\usepackage{array}     
\usepackage{subcaption}
\AtBeginDocument{%
  }

\setcopyright{acmlicensed}
\copyrightyear{2025}
\acmYear{2025}
\setcopyright{acmlicensed}
\acmConference[MM '25]{Proceedings of the 33rd ACM International Conference on Multimedia}{October 27--31, 2025}{Dublin, Ireland}
\acmBooktitle{Proceedings of the 33rd ACM International Conference on Multimedia (MM '25), October 27--31, 2025, Dublin, Ireland}
\acmDOI{10.1145/3746027.3754493}
\acmISBN{979-8-4007-2035-2/2025/10}

\begin{document}

\title{Separate Motion from Appearance: Customizing Motion via Customizing Text-to-Video Diffusion Models}


\author{Huijie Liu}
\email{liuhuijie6410@gmail.com}
\affiliation{%
  \institution{Beihang University}
  \city{Beijing}
  \country{China}
}

\author{Jingyun Wang}
\email{19231136@buaa.edu.cn}
\affiliation{%
  \institution{Beihang University}
  \city{Beijing}
  \country{China}
}

\author{Shuai Ma}
\email{msbuaa@buaa.edu.cn}
\affiliation{%
  \institution{Beihang University}
  \city{Beijing}
  \country{China}
}

\author{Jie Hu}
\email{hujiemr@163.com}
\affiliation{%
  \institution{Meituan}
  \city{Beijing}
  \country{China}
}

\author{Xiaoming Wei}
\email{hujiemr@163.com}
\affiliation{%
  \institution{Meituan}
  \city{Beijing}
  \country{China}
}
\author{Guoliang Kang}
\authornote{Corresponding author.}
\email{kgl.prml@gmail.com}
\affiliation{%
  \institution{Beihang University}
  \city{Beijing}
  \country{China}
}


\begin{abstract}
Motion customization aims to adapt the diffusion model (DM) to generate videos with the motion specified by a set of video clips with the same motion concept. 
To realize this goal, the adaptation of DM should be possible to model the specified motion concept, without compromising the ability to generate diverse appearances.
Thus, the key to solving this problem lies in how to separate the motion concept from the appearance in the adaptation process of DM.
Typical previous works explore different ways to represent and insert a motion concept into large-scale pre-trained text-to-video diffusion models, e.g., learning a motion LoRA, using latent noise residuals, etc.
While those methods can encode the motion concept, they also inevitably encode the appearance in reference videos, resulting in weakened appearance generation capability.
In this paper, we follow the typical way to learn a motion LoRA to encode the motion concept, but propose two novel strategies to enhance motion-appearance separation, including temporal attention purification (TAP) and appearance highway (AH). 
Specifically, we assume that in the temporal attention module, the pretrained Value embeddings are sufficient to serve as basic components needed by producing a new motion. 
Thus, in TAP, we choose only to reshape the temporal attention with motion LoRAs so that Value embeddings can be reorganized to produce a new motion.
Further, in AH, we alter the starting point of each skip connection in U-Net from the output of each temporal attention module to the output of each spatial attention module. 
Extensive experiments demonstrate that compared to previous works, our method can generate videos with appearance more aligned with the text descriptions and motion more consistent with the reference videos. Our homepage can be found in https://github.com/LiuHuijie6410/SeperateMotionFromAppearance.
\end{abstract}


\received{20 February 2007}
\received[revised]{12 March 2009}
\received[accepted]{5 June 2009}

\maketitle
\begin{figure}[tbp]
  \centering
  \includegraphics[width=1.0\linewidth]{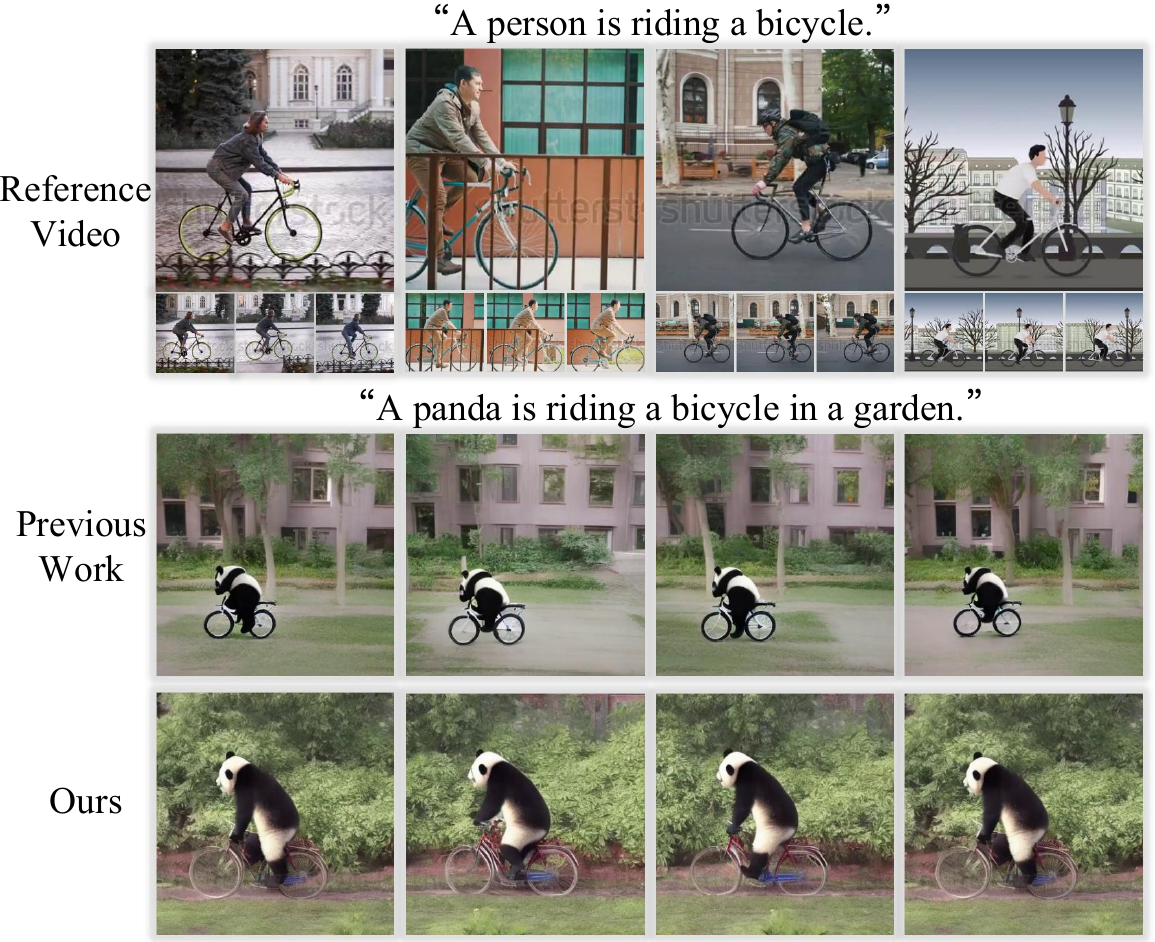}
  \Description{Limitations of previous methods.}
   \caption{\textbf{Limitations of previous methods.} Previous methods~\cite{zhao2023motiondirector} suffer from the issue of ``appearance leakage'', where elements from the reference videos, such as ``window'', unexpectedly appear in the generated video.}
   \label{fig:intro}
\end{figure}
\section{Introduction}
\label{sec:intro}
Text-to-video diffusion models~\cite{videoworldsimulators2024,blattmann2023stable} have made remarkable progress. 
With a text input, these models can generate high-quality videos that faithfully reflect the provided prompt. 
However, relying solely on text prompts, pre-trained text-to-video diffusion models struggle to generate high-fidelity videos for highly customized motion concepts. 
To address this issue, researchers explored various ways to realize motion customization of text-to-video diffusion models. 

Formally, motion customization aims to adapt the diffusion model (DM) to generate videos with the motion specified by a set of video clips (\emph{i.e.,} reference videos) with the same motion concept.
To realize this goal, the adaptation of DM should be possible to model the specified motion concept, without compromising the ability to generate diverse appearance.
Thus, the key to solving this problem lies in how to separate the motion concept from the appearance in the adaptation process of diffusion models.
Typical previous works~\cite{zhao2023motiondirector, ren2024customize,jeong2024vmc} explore different ways to represent and insert a motion concept into large-scale pre-trained text-to-video diffusion models. For example, ~\cite{zhao2023motiondirector} trains motion LoRAs to encode the motion concept from reference videos. Additionally,~\cite{zhao2023motiondirector} trains appearance LoRAs to encode the appearance of reference videos. At inference, the appearance LoRAs are abandoned to avoid overfitting to the appearance of reference videos.
Other works like~\cite{jeong2024vmc} utilize residual vectors between consecutive frames to represent a motion. 
Although these methods can encode the motion concept, there exists no explicit constraint that appearance information should not be learned or encoded into the motion representations.
Actually, they inevitably encode the appearance of reference videos (termed as ``appearance leakage''), resulting in a weakened appearance generation ability.
As shown in Fig.~\ref{fig:intro} (first row), the background of the reference videos includes some ``window'', which has nothing to do with the motion concept. Previous methods (second row), while fitting the motion, also mistakenly fit the appearance of the ``window''. 
As a result, the generated video's background also includes the ``window'', which
is not aligned with the text description. 

In this paper, we follow the typical way to learn motion LoRAs to encode the motion concept but propose two novel strategies to enhance motion-appearance separation, including temporal attention purification (TAP) and appearance highway (AH).
We choose to adopt the typical text-to-video diffusion model (T2V-DM) ZeroScope~\cite{zeroscope}
or ModelScope~\cite{wang2023modelscope} to perform motion customization.
Specifically, we assume that in the temporal transformers of text-to-video diffusion models, the pre-trained Value embeddings are sufficient to serve as basic components needed to produce a new motion. Thus, in TAP, we choose only to reshape the temporal attention with motion LoRAs so that Value embeddings can be reorganized to produce a new motion. 
As TAP may not perfectly avoid encoding appearance information in motion LoRAs,
we propose AH to maintain the appearance generation ability of original T2V-DM and further avoid appearance leakage issue. Specifically, we alter the starting point of each
skip connection in spatial-temporal U-Net~\cite{cciccek20163d,ho2022video} from the output of each temporal transformer to the output of each spatial transformers. 
The underlying logic of AH is that as the skip connection of U-Net mainly conveys high-frequency appearance information~\cite{si2023freeu}, the starting point alternation can provide a ``highway'' for the hidden states of \textit{non-adapted} spatial transformers and cut the shortcut from the hidden states of \textit{LoRA-adapted} temporal transformers. 
We empirically find that through AH, the hidden states from the decoding branch of spatial-temporal U-Net can encode appearance information more similar to the vanilla T2V-DM and motion information more similar to the T2V-DM adapted with TAP. This means AH largely maintains the appearance generation ability of vanilla T2V-DM, without harming the motion concept learning from reference videos.
With the proposed TAP and AH, our method can largely mitigate the appearance leakage issue and generate videos with customized motion and high-fidelity appearance well aligned with the text descriptions, as shown in Fig.~\ref{fig:intro} (last row).
Additionally, we point out that motion modeling occurs in the early steps of the denoising process and propose phased LoRA integration (PIL) to maximize the T2V-DM's appearance modeling capability.
Extensive experiments demonstrate that our method effectively separates motion from appearance and outperforms the previous state-of-the-art.

In a nutshell, our contributions are summarized as follows,
\begin{itemize}
    \item We carefully examine the attention mechanism of typical T2V-DMs and propose Temporal Attention Purification (TAP) to reduce the appearance information encoded in the motion LoRAs, without disturbing the motion concept learning.
    \item We propose Appearance Highway (AH) strategy, which alters the starting point of skip connection in U-Net from the hidden states of LoRA-adapted temporal transformers to those of non-adapted spatial transformers, to maintain the appearance generation ability of vanilla T2V-DM and further avoid appearance leakage issue.
    \item We conducted extensive qualitative and quantitative experiments, demonstrating that our method outperforms the previous state-of-the-art.
\end{itemize}

\begin{figure*}
  \centering
  \begin{subfigure}{0.85\linewidth}

  \includegraphics[width=\linewidth]{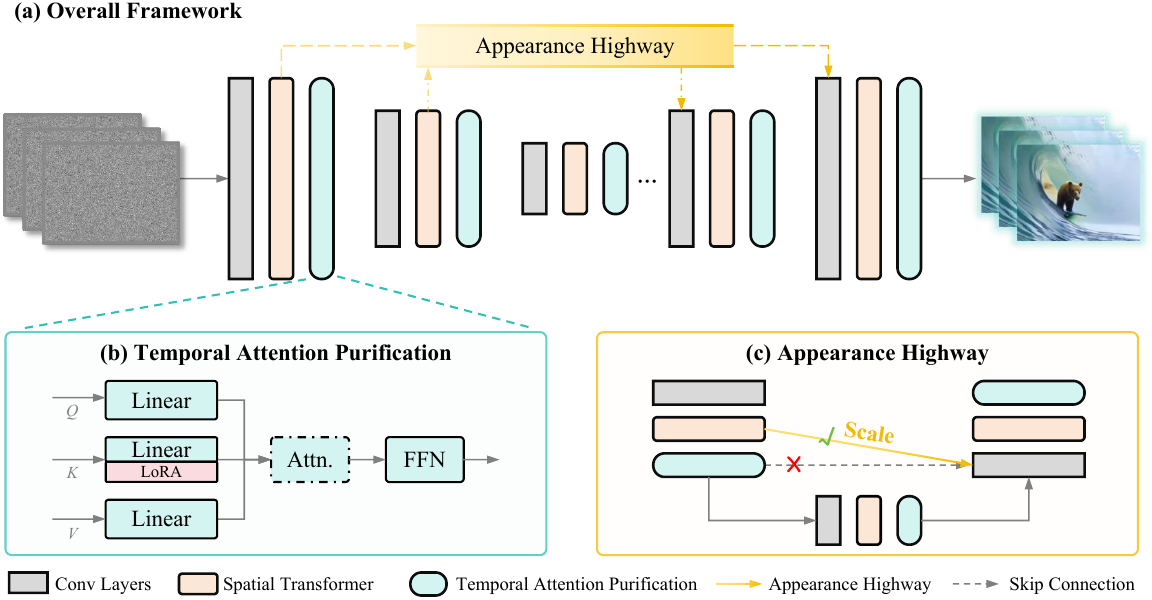}    
  \end{subfigure}
  
  \caption{Overview of our method. This figure illustrates our main contributions: (a) Overall inference process.  
  (b) Temporal Attention Purification where we utilize LoRA only to adapt the Key embeddings to reshape the temporal attention.
  (c) Appearance Highway where we alter the starting point of skip connection from LoRA-adapted temporal transformer block to non-adapted spatial transformer block.
}
\Description{Overview of our method.}
  \label{fig:overview}
  
\end{figure*}

\section{Related Work}

\label{sec:formatting}
\subsection{Text-to-Video Generation.}

Text-to-video generation is a task that involves generating realistic videos based on text prompts. Early research focused on Generative Adversarial Networks (GAN) \cite{vondrick2016generating,ijcai2019p276,tian2021goodimagegeneratorneed,shen2023mostganvvideogenerationtemporal}, autoregressive transformers\cite{yan2021videogpt,hong2022cogvideolargescalepretrainingtexttovideo,moing2021ccvscontextawarecontrollablevideo,ge2022longvideogenerationtimeagnostic}. Recently, diffusion models have emerged as the most popular paradigm~\cite{videoworldsimulators2024,an2023latent,esser2023structure,huang2024free,zhang2023show,zhang2023i2vgen,zhang2023controlvideo,blattmann2023stable}. Make-A-Video \cite{singer2022make} is first trained on labeled images and then on unlabeled videos to address the problem of the lack of paired video-text data. Imagen Video\cite{ho2022imagen} generates high-quality videos using a text-to-video model combined with spatial and temporal super-resolution models. MagicVideo\cite{zhou2022magicvideo} and VideoLDM\cite{blattmann2023align} adapt the Latent Diffusion Model's \cite{rombach2022high} architecture for video generation. In contrast to these methods, Text2Video-Zero\cite{text2video-zero}, AnimateDiff\cite{guo2023animatediff} leverage the prior knowledge of image generation models to generate videos. 
Recently, open-source T2V-DM, ModelScope~\cite{wang2023modelscope} and ZeroScope~\cite{zeroscope} have attracted considerable attention.

\subsection{Motion Customization Generation.} 

To generate motion-customized videos, several video editing methods~\cite{ceylan2023pix2video,yang2023rerender,geyer2023tokenflow,qi2023fatezero}, including Control-A-Video~\cite{chen2023control} and VideoComposer~\cite{wang2024videocomposer}, transfer motion from a reference video to new appearances. 
These approaches primarily modify the appearance without fully capturing the essence of motion, which limits appearance diversity.
Tune-A-Video~\cite{wu2023tune} extends text-to-image models into the video domain, enabling customized video generation. VMC~\cite{jeong2024vmc} uses noise residuals to represent motion information and minimizes residual loss to fine-tune pre-trained text-to-video models. LAMP~\cite{wu2024lamp} and DreamVideo~\cite{wei2024dreamvideo} incorporate an additional image paired with editing text as input, serving as guidance for the generation process. However, this approach increases the complexity of user input requirements.
To address these challenges, MotionDirector~\cite{zhao2023motiondirector} and Customize-a-Video~\cite{ren2024customize} introduce prior knowledge of video appearance into their models, encouraging a stronger focus on learning motion rather than appearance.

%

\section{Method}

%
In this paper, we aim to adapt the text-to-video diffusion model (T2V-DM) to perform motion customization, \emph{i.e.,} we need to adapt T2V-DM with reference videos containing one specific motion concept to generate new high-fidelity videos with the same motion concept.

The overview of our method is illustrated in Fig.~\ref{fig:overview}.
Our method is based on typical pre-trained T2V-DMs (\emph{e.g.,} ZeroScope~\cite{zeroscope} or ModelScope~\cite{wang2023modelscope}) whose architecture is a 3D U-Net. The basic components for such a 3D U-Net include spatial transformer and temporal transformer, which encode the spatial and temporal information, respectively.
Skip connections are utilized to connect the encoders and decoders of the U-Net.

Following MotionDirector, we adopt a dual-path LoRA adaption for training (Sec.~\ref{3.1baseline}), which tunes the model with low-rank adaptions (LoRAs) while keeping the pre-trained parameters fixed, and finally models the motion information in the LoRAs of temporal transformers. 
However, such a strategy still inevitably encodes the appearance of the reference videos into temporal LoRAs, resulting in a weakened appearance generation ability.
Thus, we propose two effective modifications.
First, instead of injecting LoRAs into all linear layers of the attention mechanism in temporal transformers, we propose a Temporal Attention Purification (TAP) strategy (Sec.~\ref{3.2purified}).
Secondly, at the inference stage, we modify the skip connection between the encoder and decoder of spatial-temporal U-Net with an Appearance Highway strategy (Sec.~\ref{3.3highway}), which transfers the hidden states from the spatial transformers rather than the temporal transformers of the encoder to the decoder.
Finally, during inference, we adopt a phased LoRA integration trick where we utilize our modified framework to denoise for a few steps and then use the original spatial-temporal U-Net for the latter steps (Sec.~\ref{3.4phased}).

\subsection{Baseline: Dual-Path LoRA Adaption}

\label{3.1baseline}
In this paper, we adopt MotionDirector~\cite{zhao2023motiondirector} as our baseline. 
MotionDirector utilizes a spatial-temporal U-Net~\cite{cciccek20163d,ho2022video}, which takes a visual latent code $z_t\in \mathcal{R}^{b \times f\times h\times w \times c}$ ($b$, $f$, $h$, $w$, $c$ representing batch size, the number of frame, height, width, and
channel dimensions, respectively) at each time step $t$ and textual feature $y$ as input.
MotionDirector adapts ZeroScope~\cite{zeroscope} to perform motion customization.

MotionDirector adopts a dual-path LoRA adaption:
For the spatial path, MotionDirector injects spatial LoRAs into self-attention layers of spatial transformers to reconstruct the appearance of training data, supervised by a random frame from the reference video.
For the temporal path, MotionDirector keeps the spatial LoRAs frozen while injecting temporal LoRAs into temporal transformers to model the motion information with the following loss, supervised by the entire reference videos: 
\begin{equation}
    \min_{\theta_{tp}}\mathbb{E}_{z_0, y, t, \epsilon} 
\left[ \left\lVert \epsilon - \epsilon_{\theta,\theta_{tp}}(z_t, y, t) \right\rVert_2^2 \right],
\label{eq1}
\end{equation}
where $\epsilon$ denotes the noise added in $z_t$ and $\epsilon_{\theta,\theta_{tp}}$ denotes the U-Net $\epsilon_\theta$ with temporal LoRAs $\theta_{tp}$. 
As a result, MotionDirector models the motion concept within the trained temporal LoRAs. At inference, it generates videos simply utilizing trained temporal LoRAs. 

However, we observe that MotionDirector tends to overfit the appearance of the reference videos, suggesting that the trained temporal LoRAs still encode the appearance information. It is reasonable as with explicit regularization, the dual-path LoRA adaptation framework cannot guarantee the temporal LoRA does not encode the appearance information.
In this paper, we carefully examine the architecture of typical T2V-DMs and propose two novel strategies to constrain temporal LoRA learning, including temporal attention purification (TAP) and appearance highway (AH).


\subsection{Temporal Attention Purification}

\label{3.2purified}

In this section, we propose Temporal Attention Purification strategy (TAP) to separate motion encoding from appearance encoding in adapting the temporal transformer block with temporal LoRAs.

Specifically, we assume that in the temporal attention module, the pre-trained Value embeddings ($W_V$) are sufficient to serve as the basic components needed for depicting a motion. 
To generate a new motion, what we need to adapt is the temporal attention, which determines how we combine the Value embeddings in the temporal attention module.
Thus, in TAP, we choose to utilize temporal LoRAs only to adapt the Query ($W_Q$) or the Key embeddings ($W_K$) to reshape the temporal attention to produce a new motion, without disturbing the Value embeddings.
As the temporal attention is only utilized to combine the Value embeddings, we expect reshaping it will not introduce appearance information.


We probe each component of the temporal transformer block to verify our assumption.
Specifically, we evaluate the effects of adapting different parts of a temporal transformer block.
By adapting different parts of the temporal transformer with LoRAs, we have different results. 
We utilize an off-the-shelf motion classifier~\cite{Wu_2023_CVPR} to classify the generated videos.
Higher accuracy (denoted as ``motion'' in Tab.~\ref{tab:purify}) mean higher quality of the generated motions.
We remove the verbs from the edited text prompt and calculate the CLIP scores between the verb-free text prompt and the video as a metric of appearance alignment (App. Align.). 
We then calculate the CLIP scores between text ``a person'' and the video as a metric of appearance leakage (App. Leak). 
For example, we fine-tune the model using several reference videos of a person playing golf, then generate a video with the prompt ``a monkey is playing golf on a field full of flowers.''
To evaluate appearance alignment, we compute the CLIP similarity between the generated video and the text ``a monkey is on a field of flowers''.
To measure appearance leakage, we compute the CLIP similarity using the text ``a person.''

Higher ``App. Align.'' and lower ``App. Leak.'' means better appearance quality of the generated videos.
From the Tab.~\ref{tab:purify}, we observe that compared to the T2V-DM, adapting Query, Key, Value and the feed-forward network ($W_{ff}$) of temporal transformer (\emph{i.e.,} MotionDirector) does improve the motion quality, but scarifies the appearance quality.
Adapting Value and the feed-forward network also harm the appearance modeling.
In contrast, reshaping the temporal attention (including reshaping Query, Key or both) yields high-quality videos considering both motion quality and appearance quality.
All those results align well with our assumption.

\begin{table}[t]
 \centering
 \caption{Probing experiments on the temporal attention transformer block. For metric definitions, refer to Sec.~\ref{3.2purified}}
 
 \resizebox{0.44\textwidth}{!}{ 
 \begin{tabular}{@{}lccc@{}}
   \toprule
   Module & Motion $\uparrow$ & App. Align. $\uparrow$ & App. Leak. $\downarrow$ \\
   \midrule
   T2V-DM & 46.8 & 26.38 & \textbf{20.82}\\
   MotionDirector & 64.1 & 21.26 & 21.66\\
   $W_Q$ & 64.1 & 25.51 & 21.33\\
   $W_K$ & 64.3 & \textbf{25.52} & 21.33\\
   $W_V$ & 62.4 & 22.88 & 21.46\\
   $W_{ff}$ & \textbf{69.2} & 21.23 & 21.83\\
   $W_Q,W_K$ & 63.9 & 25.48 & 21.35 \\
   \bottomrule
 \end{tabular}
 }
 \label{tab:purify}
 \end{table}
  
Based on our observations, we choose to only adapt the Key embedding of a temporal transformer block with temporal LoRAs. 


\subsection{Appearance Highway} 

\label{3.3highway}
\begin{figure}
  \centering  \includegraphics[width=\linewidth]{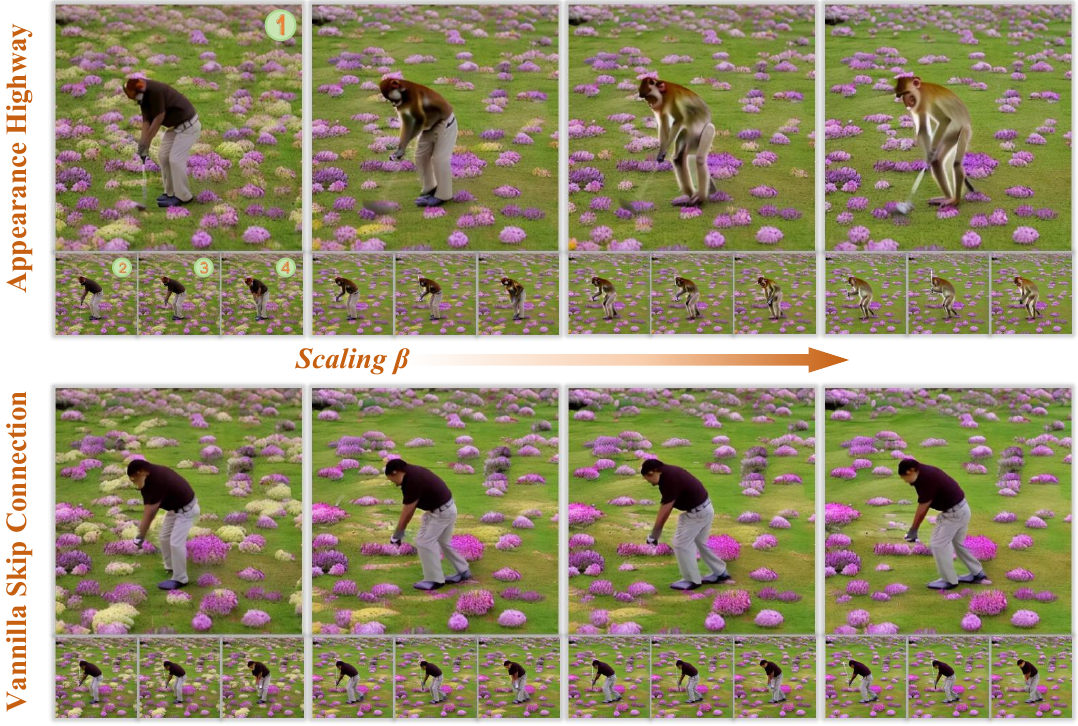}    
  
  \caption{\textbf{The comparison between the Appearance Highway and the Skip Connection is illustrated in the figure.} The figure presents videos generated by ``A monkey is playing golf on a field full of flowers''. The first row shows that as the scale of the appearance highway increases, appearance leakage is progressively alleviated. In contrast, the second row shows that increasing the scale of the vanilla skip connection has no impact on appearance leakage. The smaller images below each main image represent subsequent frames of the video.}
  \Description{The comparison between the Appearance Highway and the Skip Connection is illustrated in the figure.}
  \label{fig:highway}
\end{figure}

As TAP may not perfectly avoid encoding appearance information in motion LoRAs, we
propose using Appearance Highway (AH) strategy to maintain the appearance generation ability of the original T2V-DM and further avoid appearance leakage issues.
Specifically, we alter the starting point of each
skip connection in spatial-temporal U-Net~\cite{cciccek20163d,ho2022video} from the output of each \textit{temporal} transformer to the output of each \textit{spatial} transformer. 
In other words, the appearance highway is an improved skip connection designed to decouple motion and appearance.
The starting point alternation can provide a ``highway'' for the hidden-states of \textit{non-adapted} spatial transformers and cut the shortcut from the hidden-states of \textit{LoRA-adapted} temporal transformers.
As shown in Fig.~\ref{fig:highway}, the videos are generated conditioned on ``A monkey is playing golf on a field full of flowers''. 
The first row of Fig.~\ref{fig:highway} illustrates the videos generated with AH. We further show how the appearance changes as we enhance the effect of hidden states from the spatial transformers by varying scales. 
Compared with adapted T2V-DM with vanilla skip connection, AH yields videos with appearance more aligned with text of the target.
Moreover, as the effect of hidden states from the spatial transformers is enhanced by increasing scales (see the first row of Fig.~\ref{fig:highway}), the appearance of generated videos becomes more aligned with the text descriptions.
In contrast, with vanilla skip connection, no matter how we enhance the temporal transformer's output, the appearance cannot match the text.

\begin{figure}[t]
  \centering
  \includegraphics[width=\linewidth]{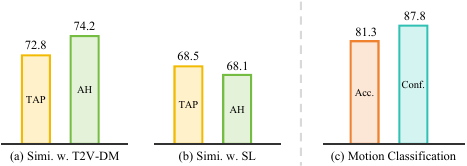} 
   \caption{Investigation of AH. (a) shows the similarity between the TAP/AH's hidden states and T2V-DM's. (b) shows the similarity between the hidden states of TAP/AH and hidden states of T2V-DM with inserted spatial LoRAs. (c) shows the accuracy and confidence of the motion classifier.}
   \label{fig:similarity_highway}
\Description{Investigation of AH.}   
\end{figure}

Furthermore, we find that AH may largely maintain the appearance generation ability of vanilla T2V-DM, without harming the motion concept learning from the reference videos. 
We conducted experiments to verify this.
As changing the starting point of skip connection directly affects the decoding results, we extract the hidden states from the decoding part after projecting the hidden states combined from the starting point of skip connection and the decoder input (the pink part illustrated in Fig.~\ref{fig:overview}(c)).

We compare hidden states from 4 kinds of models: (1) T2V-DM with high appearance generation ability; (2) T2V-DM with spatial-LoRA (SL) which customizes the appearance of reference videos; (3) T2V-DM with TAP
(TAP); (4) TAP combined with AH (AH).

As shown in Fig.~\ref{fig:similarity_highway} (a), we compare the similarity between AH and T2V-DM and that between TAP and T2V-DM. We observe that with AH, the hidden states of decoder are more similar to T2V-DM, which indicates AH may further improve the appearance generation ability of adapted T2V-DM. 
We also compare the similarity between AH and SL and that between TAP and SL in Fig.~\ref{fig:similarity_highway} (b). We observe that with AH, the hidden states of the decoder are more dissimilar to SL compared to TAP.
All of these results indicate that AH may enhance the modeling ability of the appearance beyond TAP.

We extract optical flows from TAP's hidden states as positive samples and those from T2V-DM's hidden states as negative samples to train a motion classifier. 
Then we use this classifier to evaluate the hidden states from AH. 
Fig.~\ref{fig:similarity_highway} shows most hidden states from AH can be classified into TAP rather than vanilla T2V-DM in a high accuracy and with high confidence.
These results show that utilizing AH may not disturb the motion generation ability of TAP.

We multiply the starting point of AH, \emph{i.e.,} hidden states of spatial transformer by a constant $\beta$ to strengthen its effect. AH can be used during both training and inference, or exclusively during inference. Our experiments show that AH utilized during training and AH utilized during inference yield comparable results. For simplicity, we only use it during the inference process (see Sec.~\ref{sec:exp}). 

\subsection{Phased LoRA Integration}

\begin{figure*}
  \centering
    \includegraphics[width=\linewidth]{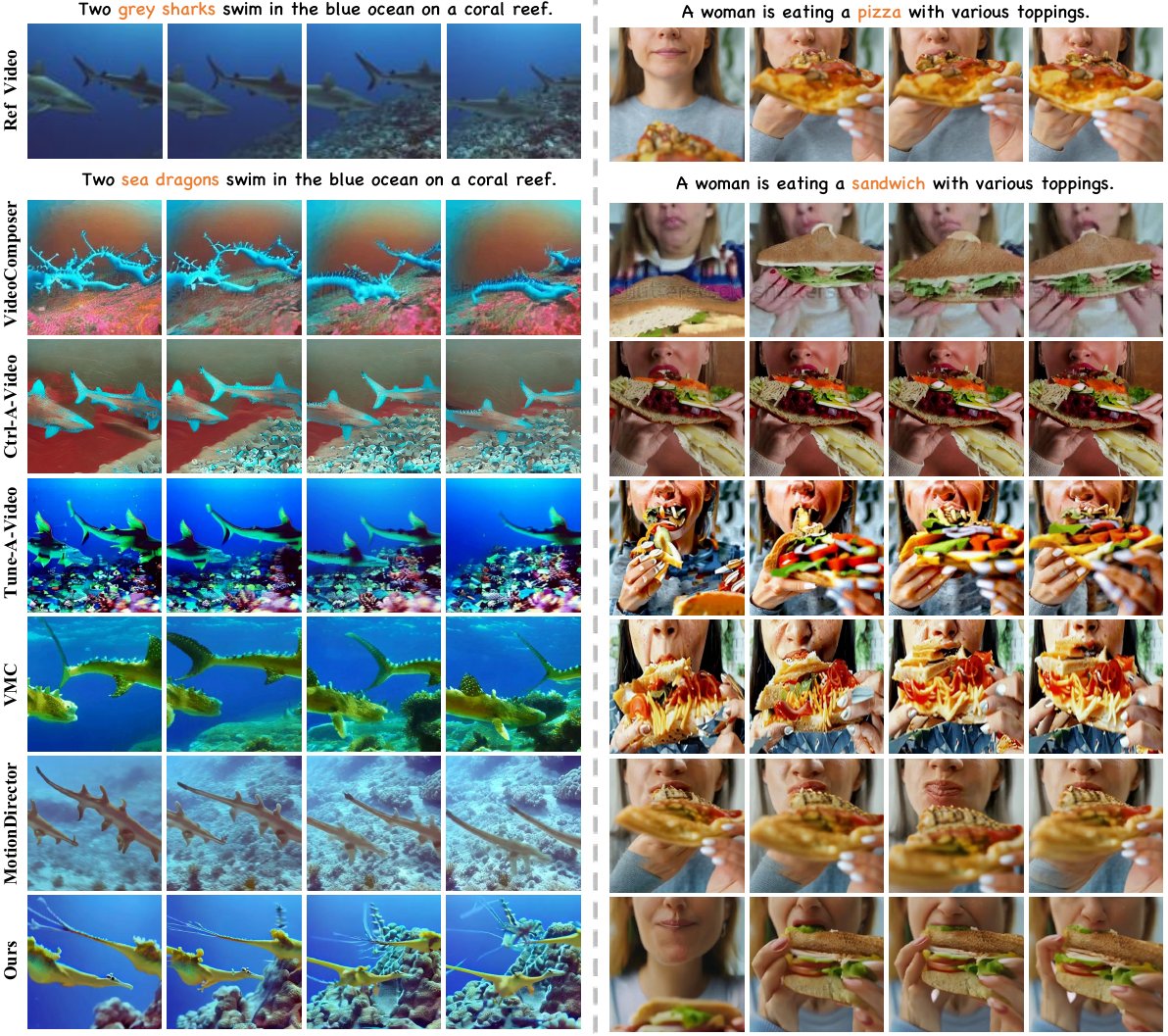} 
    
    \caption{\textbf{Qualitative results on one-shot.} 
    Tune-A-Video~\cite{wu2023tune}, VMC~\cite{jeong2024vmc}, MotionDirector~\cite{zhao2023motiondirector} and our method need to fine-tune. VideoComposer~\cite{wang2024videocomposer} and Ctrl-A-Video~\cite{chen2023control} are training free.}
    \label{fig:one-shot}
  \hfill
\Description{Qualitative results on one-shot.}
\end{figure*}

\begin{table*}[htbp]
\caption{\textbf{Quantitative results on few-shot.} In human evaluation, the paired numbers show our method’s voting rate on the left, while previous methods' voting rate on the right.}
  \centering
  \small
\begin{tabularx}{0.98\textwidth}{@{} 
    >{\raggedright\arraybackslash}m{2.3cm}  
    >{\centering\arraybackslash}p{1.45cm}  
    >{\centering\arraybackslash}p{1.45cm}  
    >{\centering\arraybackslash}p{1.45cm}  
    >{\centering\arraybackslash}p{1.45cm}  
    >{\centering\arraybackslash}p{1.45cm}  
    >{\centering\arraybackslash}p{1.6cm}  
    >{\centering\arraybackslash}p{1.6cm}  
    >{\centering\arraybackslash}p{1.6cm}  
    @{}}
    \toprule
    \multicolumn{1}{c}{} & 
    \multicolumn{5}{c}{Automatic Evaluations} & 
    \multicolumn{3}{c}{Human Evaluations (Ours v.s. Others)} \\
    \cmidrule(lr){2-6} \cmidrule(lr){7-9}
    \multicolumn{1}{l}{Methods} & \makecell[c]{Text \\ Align. $\uparrow$} & \makecell[c]{Temporal \\ Consist. $\uparrow$} & \makecell[c]{Aesthetic\\Score $\uparrow$} & \makecell[c]{ViCLIP \\ Score $\uparrow$} & \makecell[c]{VBench\\Average $\uparrow$} &\makecell[c]{Appearance \\ Diversity}  & \makecell[c]{Motion \\ Fidelity}  & \makecell[c]{Temporal \\ Consistency}  \\
    \midrule
        VideoComposer~\cite{wang2024videocomposer} & 27.34 & 90.00 & 20.29 & 25.63 & 74.12 &65.5/34.5 & 63.6/36.4 & 67.3/32.7\\
        Control-A-Video~\cite{chen2023control} & 25.56 & 93.40 & 19.80 & 23.66 & 73.33& 74.5/25.5 & 58.2/41.8 & 58.2/41.8\\
        Tune-A-Video~\cite{wu2023tune}& 25.69 & 92.11 & 20.18 & 23.37 & 75.34 &58.2/41.8 & 56.4/43.6 & 60.0/40.0\\
        VMC~\cite{jeong2024vmc}& 25.67 & 90.50 & 20.46 & 23.98 & 74.23& 81.8/19.2 & 69.1/30.9 & 83.6/16.4\\
        MotionDirector~\cite{zhao2023motiondirector}& 27.55 & 93.61 & 20.36 & 25.54 & 75.95& 61.8/38.2 & 58.2/41.8 & 63.6/36.4\\
        Ours & \textbf{28.52} & \textbf{93.83} & \textbf{20.55} &\textbf{26.52} & \textbf{76.33} & - & - & - \\
    \bottomrule
  \end{tabularx}
  \label{tab:oneshot2}
\end{table*}


\label{3.4phased}
\begin{figure*}[ht]
  \centering
  \includegraphics[width=\linewidth]{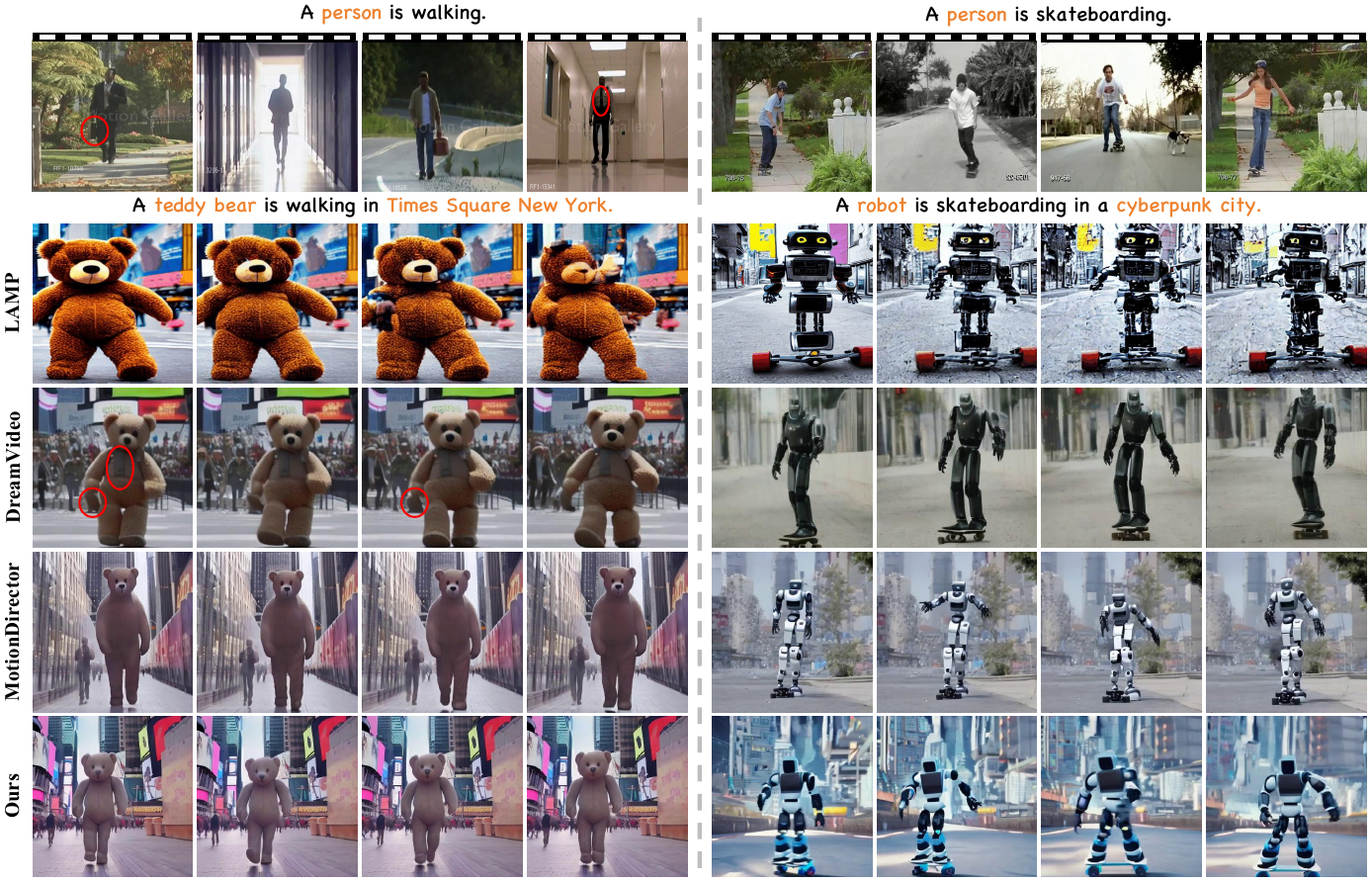}  
    \caption{\textbf{Qualitative results on few-shot.} During training, few videos from UCF Sports dataset~\cite{rodriguez2008action,soomro2015action} are used as reference videos. During inference, we use Stable Diffusion to generate images paired with the texts, which serve as the image conditions for LAMP~\cite{wu2024lamp} and DreamVideo~\cite{wei2024dreamvideo}. MotionDirector~\cite{zhao2023motiondirector} and our method do not need image conditions.}
  \hfill
\Description{Qualitative results on few-shot.}
  \label{fig:few-shot}
\end{figure*}

\begin{table*}[htbp]
\caption{\textbf{Quantitative results on few-shot.} In human evaluation, the paired numbers show our method’s voting rate on the left, while previous methods' voting rate on the right.}
  \centering
  \small
\begin{tabularx}{0.95\textwidth}{@{} 
    >{\raggedright\arraybackslash}m{2.1cm}  
    >{\centering\arraybackslash}p{1.45cm}  
    >{\centering\arraybackslash}p{1.45cm}  
    >{\centering\arraybackslash}p{1.45cm}  
    >{\centering\arraybackslash}p{1.45cm}  
    >{\centering\arraybackslash}p{1.45cm}  
    >{\centering\arraybackslash}p{1.6cm}  
    >{\centering\arraybackslash}p{1.6cm}  
    >{\centering\arraybackslash}p{1.6cm}  
    @{}}
    \toprule
    \multicolumn{1}{c}{} & 
    \multicolumn{5}{c}{Automatic Evaluations} & 
    \multicolumn{3}{c}{Human Evaluations (Ours v.s. Others)} \\
    \cmidrule(lr){2-6} \cmidrule(lr){7-9}
    \multicolumn{1}{l}{Methods} & \makecell[c]{Text \\ Align. $\uparrow$} & \makecell[c]{Temporal \\ Consist. $\uparrow$} & \makecell[c]{Aesthetic\\Score $\uparrow$} & \makecell[c]{ViCLIP \\ Score $\uparrow$} & \makecell[c]{VBench\\Average $\uparrow$} &\makecell[c]{Appearance \\ Diversity}  & \makecell[c]{Motion \\ Fidelity}  & \makecell[c]{Temporal \\ Consistency}  \\
    \midrule
        LAMP~\cite{wu2024lamp} & 26.19 & 87.89 & 18.41 & 22.68 & 67.41& 59.1/40.9 & 63.6/36.4 & 60.6/39.4 \\ 
        DreamVideo~\cite{wei2024dreamvideo} & 27.48 & 90.52 & 19.17 & 25.10 & 71.75& 66.7/33.3 & 53.0/47.0 & 50.0/50.0\\ 
        MotionDirector~\cite{zhao2023motiondirector}& 29.74 & 94.27 & 19.93 & 27.83 & 72.91& 59.1/40.9 & 51.5/48.5 & 48.5/51.5 \\
        Ours & \textbf{29.89} & \textbf{94.92} & \textbf{19.95} & \textbf{28.53} & \textbf{74.17} & - & - & - \\
    \bottomrule
  \end{tabularx}
  \label{tab:fewshot2}
\end{table*}

We find that as denoising proceeds, the model tends to learn more fine-grained appearance information and the risk of overfitting to the appearance of the reference videos (appearance leakage) becomes higher. Thus, we propose utilizing a phased LoRA integration strategy during inference to mitigate the risk of appearance leakage. Specifically, in the early steps of denoising, we utilize T2V-DM adapted with TAP and AH, and in the remaining steps of denoising, we utilize the vanilla T2V-DM. The formula of phased LoRA integration is 
\begin{equation}
{z}_{t-1} = 
\begin{cases}
    \frac{1}{\sqrt{\alpha_t}} \left({z}_t - \frac{1 - \alpha_t}{\sqrt{1 - \bar{\alpha}_t}} \, \epsilon_\theta({z}_t, t) \right) + \sigma_t, & t \le \tau \\
    \frac{1}{\sqrt{\alpha_t}} \left({z}_t - \frac{1 - \alpha_t}{\sqrt{1 - \bar{\alpha}_t}} \, \epsilon_{\theta,\theta_{K}}({z}_t, t) \right) + \sigma_t, & t > \tau
\end{cases}
\end{equation}
where $\alpha$ is a constant controlling the noise strength.

%
\section{Experiment}

\label{sec:exp}
\subsection{Experimental Setup}

\textbf{Datasets.}
For evaluations on one video, we utilize LOVEU-TGVE-2023 dataset~\cite{wu2023cvpr}. The dataset includes 76 reference videos, each accompanied by 4 prompts. Following ~\cite{zhao2023motiondirector}, we design 3 additional prompts with significant appearance changes for each video to evaluate the diversity of the outputs. 
For evaluations on few-shot videos, we select 3-5 reference videos for each of the 13 categories in the UCF Sports dataset~\cite{rodriguez2008action,soomro2015action}.
We design 8 prompts for the corresponding reference videos for each category. 
Similar to MotionDirector, we employ a large language model to create diverse prompts, each composed of [\#Object] + [\#Motion] + [\#Appearance]. While the [\#Object] remains consistent with the original prompts (for UCF data, [\#Object] is the category name), the [\#Motion] and [\#Appearance] are very diverse.

\textbf{Baselines}
For evaluations on one-shot, we compare our method with the popular motion customization method, including MotionDirector~\cite{zhao2023motiondirector} and VMC~\cite{jeong2024vmc}.
In addition, we compare with controllable video generation methods, including VideoComposer~\cite{wang2024videocomposer}, Control-A-Video~\cite{chen2023control}, and Tune-a-Video~\cite{wu2023tune}. Tune-a-Video, VMC, MotionDirector, and our method need fine-tuning, whereas VideoComposer and Control-A-Video do not need training.
For evaluations on few-shot, we compare our approach with DreamVideo~\cite{wei2024dreamvideo}, LAMP~\cite{wu2024lamp}, and MotionDirector. 

\textbf{Metrics}
For automatic evaluations, we use three metrics: text alignment (Text Align.), temporal consistency (Temporal Consist.), and aesthetic score. 
Text alignment measures the average CLIP similarity~\cite{radford2021learning} between each and the text prompt, reflecting how well the generated video aligns with the given description. 
Temporal consistency assesses the video's coherence by CLIP similarity between frames. The aesthetic score is calculated using an aesthetic model~\cite{Kirstain2023PickaPicAO}.
To provide a more comprehensive evaluation, we also leverage ViCLIP~\cite{wang2023internvid} and VBench~\cite{huang2023vbench}. ViCLIP is used as a supplementary metric alongside TextCLIP. 
VBench is a benchmark that evaluates models across multiple dimensions. 
Due to space limitations, we report the average of these results (VBench Average), with detailed results provided in the appendix.
For human evaluations, we invite 11 participants to complete a survey, resulting in a total of 1,419 responses.
The survey assesses appearance diversity, motion fidelity, and temporal consistency. We show two videos: one generated by our method and the other by previous methods. For each evaluation dimension, users vote for the better video.

\subsection{Evaluations on One-Shot}
\textbf{Quantitative results.}
For automatic evaluations in Tab.~\ref{tab:oneshot2}, our method significantly improves text alignment and ViCLIP score, demonstrating that it suppresses appearance leakage.
Our method also demonstrates superiority in other dimensions.

\textbf{Qualitative results.}
In the example on the left of Fig.~\ref{fig:one-shot}, 
previous methods generate ``grey sharks'' exhibiting overfitting issue, while our method generates a realistic ``sea dragon''. 
In the example on the right, our method generates a vivid ``sandwich'', while the ``sandwich'' produced by other methods resembles the ``pizza'' from the reference video.

\subsection{Evaluation on Few-Shot}

\textbf{Quantitative results.}
As shown in Tab.~\ref{tab:fewshot2}, the automatic evaluation shows our method surpasses previous approaches across multiple dimensions. 
Human evaluations show that while our temporal consistency may not surpass that of DreamVideo and MotionDirector, our appearance diversity and motion fidelity are superior.

\textbf{Qualitative Results.}
Fig.~\ref{fig:few-shot} illustrates that MotionDirector occasionally exhibits appearance leakage from the training set. In the example on the left, DreamVideo generates a teddy bear holding an object resembling a suitcase and wearing a tie. It occurs because the training set includes humans carrying suitcases and wearing ties. MotionDirector generates a background featuring vertical reflective structures, similar to the background in the second reference video.
In the right case, the backgrounds generated by previous methods lack a distinctly ``cyberpunk'' aesthetic and instead resemble those of reference videos. 
In contrast, our method avoids this issue.
\subsection{Ablation}
\textbf{Effect of Our Sub-Contributions.}
In Tab.~\ref{tab:ablation_sub}, we evaluate the capabilities of each contribution: temporal attention purification (TAP), appearance highway (AH) and phased LoRA integration (PLI). TAP, AH, and PLI each significantly improve text alignment. AH improves text alignment and aesthetic score, but slightly compromises temporal consistency. However, combined with PLI, it not only further enhances text alignment and aesthetic score, but also mitigates the loss of motion consistency caused by AH.
\begin{table}[ht]
  \centering
\caption{\textbf{Effect of each component and effect of motion customization with different T2V-DM.} We test on two popular T2V-DM models: ZeroScope(ZS) and ModelScope(MS).}
  \resizebox{0.48\textwidth}{!}{
  \begin{tabular}{@{}lcccp{11mm}cccc@{}}
    \toprule
    Models & \makecell[c]{TAP} & \makecell[c]{AH} & \makecell[c]{PLI} & \makecell[c]{Text \\ Align. $\uparrow$} & \makecell[c]{Temporal \\ Consist. $\uparrow$} & \makecell[c]{Aesthetic\\Score $\uparrow$} & \makecell[c]{ViCLIP \\ Score $\uparrow$} & \makecell[c]{VBench\\Average $\uparrow$}\\ 
    \midrule
    ZS & & & & 27.55 & 93.61 & 20.36 & 25.54 & 75.95 \\
    ZS & \checkmark & & & 27.91 & \textbf{93.86} & 20.44 & 25.64 & 76.11\\
    ZS & \checkmark & \checkmark & & 28.32 & 93.71 & 20.54 & 26.24 & 75.93\\
    ZS & \checkmark & \checkmark & \checkmark & \textbf{28.52} & 93.83 & \textbf{20.55} & \textbf{26.52} & \textbf{76.33} \\
    \midrule
    MS & & & & 26.40 & 92.19 & 19.73 & 23.10 & 74.45\\
    MS & \checkmark & & & 26.43 & 92.04 & 19.66 & 23.12 & 74.77\\
    MS & \checkmark & \checkmark & & 27.07 & 92.68 & 19.81 & 23.46 & 74.33\\
    MS & \checkmark & \checkmark & \checkmark & \textbf{27.48} & \textbf{92.72} & \textbf{19.92} & \textbf{24.20} & \textbf{75.78}\\ 
    \bottomrule
  \end{tabular}
  }
  \label{tab:ablation_sub}
\end{table}

\textbf{Effect of motion customization with different T2V-DM.} 
Tab.~\ref{tab:ablation_sub} demonstrates that our method is effective across various foundation models using a spatial-temporal U-Net, such as ZeroScope (ZS)~\cite{zeroscope} and ModelScope (MS)~\cite{wang2023modelscope}.

\textbf{Training with AH vs. Post-processing with AH.} The AH can be used during both training and inference, or exclusively during inference. As shown in Tab.~\ref{tab:ablation_traning_or_infer}, the results are comparable. For simplicity, we use it with post-processing as our method.
\begin{table}[h]
  \centering
  \caption{\textbf{Training with AH vs. Post-processing with AH:}  (1) Training: apply it during both training and inference. (2) Post-processing: during training, the vanilla skip connection is used, while at inference, AH is employed. Experiments show that these two strategies are comparable.}
  \resizebox{0.46\textwidth}{!}{
  \begin{tabular}{@{}lccccc@{}}
    \toprule
    Method & \makecell[c]{Text \\ Align. $\uparrow$} & \makecell[c]{Temporal \\ Consist.} & \makecell[c]{Aesthetic\\Score $\uparrow$} & \makecell[c]{ViCLIP \\ Score $\uparrow$} & \makecell[c]{VBench\\Average $\uparrow$}\\
    \midrule
    Training & \textbf{28.64} & 93.65 & 20.53 & \textbf{26.52} & 76.21 \\
    Post-processing & 28.52 & \textbf{93.83} & \textbf{20.55} & \textbf{26.52} & \textbf{76.33} \\
    \bottomrule
  \end{tabular}
  }

  \label{tab:ablation_traning_or_infer}
\end{table}

\textbf{Different choices of appearance highway's scale.}
We scale the starting point of AH, \emph{i.e.,} hidden states of spatial transformer by a constant $\beta$ to strengthen its effect. 
As shown in Tab.~\ref{tab:ablation_scale}, we change the $\beta$ and evaluate the model's capabilities. 
When $\beta$ is small, increasing $\beta$ enhances the appearance information generated by T2V-DM and improves text alignment and aesthetic score. 
However, when $\beta$ becomes larger, further increases in $\beta$ lead to a decrease in these results.
We hypothesize that this is due to the excessive magnitude of the original hidden states, which disrupts the data distribution and consequently degrades the generation quality.
In addition, we found that enhancing AH leads to a decrease in temporal consistency. However, we mitigate this issue through temporal attention purification and phased LoRA integration.
\begin{table}[htbp]
  \centering
  \caption{Different choices of appearance highway's scale $\beta$.}
  \resizebox{0.46\textwidth}{!}{
    \small
    \begin{tabular}{@{} 
    >{\raggedright\arraybackslash}m{0.8cm}  
    >{\centering\arraybackslash}p{1.0cm}  
    >{\centering\arraybackslash}p{1.2cm}  
    >{\centering\arraybackslash}p{1.1cm}  
    >{\centering\arraybackslash}p{1.1cm}  
    >{\centering\arraybackslash}p{1.2cm}  
    @{}}
    \toprule
    Scale & \makecell[c]{Text \\ Align. $\uparrow$} & \makecell[c]{Temporal \\ Consist. $\uparrow$} & \makecell[c]{Aesthetic\\Score $\uparrow$} & \makecell[c]{ViCLIP \\ Score $\uparrow$} & \makecell[c]{VBench\\Average $\uparrow$}\\ 
    \midrule
    $\beta$=1 & 28.13 & 93.93 & 20.52 & 26.02 & 75.72 \\
    $\beta$=1.05 & 28.14 & 93.86 & 20.53 & 26.10 & 75.87\\
    $\beta$=1.1 & 28.32 & 93.71 & 20.54 & 26.24 & 75.93\\
    $\beta$=1.15 & 28.31 & 93.60 & 20.52 & 26.20 & 76.10\\
    $\beta$=1.2 & 28.29 & 93.48 & 20.52 & 26.23 & 76.30\\
    \bottomrule
  \end{tabular}
  }
  \label{tab:ablation_scale}
\end{table}

\section{Conclusion}
\label{conclusion}
In this paper, we propose two novel strategies to separate motion from appearance during adapting T2V-DM to perform motion customization, including Temporal Attention Purification (TAP) and Appearance Highway (AH). In TAP, we utilize LoRA only to adapt the temporal attention, and in AH, we alter the starting point of skip connection from LoRA-adapted temporal transformer block to non-adapted spatial transformer block. Extensive experiments show that our method outperforms previous state-of-the-arts.

\begin{acks}
This project is supported by National Natural Science Foundation of China under Grant 92370114.
\end{acks}

\bibliographystyle{ACM-Reference-Format}
\bibliography{sample-base}

\appendix

\end{document}